\documentclass[preprint,review,10pt]{elsarticle}  %

\usepackage{amssymb}
\usepackage[numbers]{natbib}
\usepackage{amsmath}
\usepackage{booktabs}
\usepackage{makecell}
\usepackage{multirow}
\usepackage{graphicx}
\usepackage{caption}
\usepackage{hyperref}
\usepackage[vlined,ruled,commentsnumbered,linesnumbered]{algorithm2e}
\usepackage[separate-uncertainty = true,multi-part-units = repeat]{siunitx}
\usepackage{xcolor}
\usepackage{mathtools}

\DeclareMathOperator*{\argmax}{arg\,max}
\usepackage{lineno}

\journal{Pattern Recognition}

\begin{document}

\begin{frontmatter}

\title{DMT: Dynamic Mutual Training for Semi-Supervised Learning}

\author[1]{Zhengyang Feng\fnref{fn1}}
\author[1]{Qianyu Zhou\fnref{fn1}}
\author[1]{Qiqi Gu}
\author[1]{Xin Tan\fnref{fn2}}
\author[2]{Guangliang Cheng}
\author[3]{Xuequan Lu\fnref{fn2}}
\author[2]{Jianping Shi}
\author[1]{Lizhuang Ma\fnref{fn2}}
\address[1]{Shanghai Jiao Tong University, Shanghai, China}
\address[2]{SenseTime Research, Shanghai, China}
\address[3]{Deakin University, Australia}
            
\fntext[fn1]{Equal contribution.}
\fntext[fn2]{Corresponding authors. Emails: tanxin2017@sjtu.edu.cn, xuequan.lu@deakin.edu.au, ma-lz@cs.sjtu.edu.cn}

\begin{abstract}
Recent semi-supervised learning methods use pseudo supervision as core idea, especially self-training methods that generate pseudo labels. However, pseudo labels are unreliable. Self-training methods usually rely on single model prediction confidence to filter low-confidence pseudo labels, thus remaining high-confidence errors and wasting many low-confidence correct labels. In this paper, we point out it is difficult for a model to counter its own errors. Instead, leveraging inter-model disagreement between different models is a key to locate pseudo label errors. With this new viewpoint, we propose mutual training between two different models by a dynamically re-weighted loss function, called Dynamic Mutual Training (DMT). We quantify inter-model disagreement by comparing predictions from two different models to dynamically re-weight loss in training, where a larger disagreement indicates a possible error and corresponds to a lower loss value. Extensive experiments show that DMT achieves state-of-the-art performance in both image classification and semantic segmentation. Our codes are released at \url{https://github.com/voldemortX/DST-CBC}.

\end{abstract}

\begin{keyword}
dynamic mutual training \sep inter-model disagreement \sep noisy pseudo label \sep semi-supervised learning
\end{keyword}

\end{frontmatter}

\section{Introduction}
\label{sec:1}

In recent years, with the rise of deep learning, substantial improvements have been shown in various computer vision tasks, e.g. image classification \cite{zagoruyko2016wide,wu2019wider} and semantic segmentation \cite{deeplabv2,zhao2017pyramid}. However, deep learning methods require a large amount of annotated data to learn generalized representations. Although a large-scale dataset is easily gathered from cameras or web pages, the labor cost for labeling such a dataset has become unbearable in many real-world applications. For example, it takes 1.5 hours for a human annotator to label a high-resolution image of urban street scenes with pixel-wise annotations \cite{cordts2016cityscapes}.
In this work, we focus on semi-supervised learning to alleviate the label costs, by taking semantic segmentation and image classification as examples.

Semi-supervised learning labels only a small part of the dataset (labeled subset), and exploits the remaining part as unlabeled data (unlabeled subset). To learn without labels, a natural idea is  ``bootstrapping'' (pulling oneself up by one's own bootstraps) \cite{yarowsky1995unsupervised}, i.e. using self (pseudo) supervisions.
Two lines of approaches have achieved good performance on both semi-supervised image classification and semantic segmentation: entropy minimization (i.e. self-training) \cite{lee2013pseudo,hung} and consistency regularization \cite{tarvainen2017mean,french2019semisupervised}. Recently, hybrid methods that combine those two directions, MixMatch with sharpening \cite{berthelot2019mixmatch}, s4GAN + MLMT with separate network branches \cite{mittal2019semi}, show state-of-the-art performance on image classification and semantic segmentation, respectively. Our method is based on offline self-training with data augmentation as consistency regularization, a hybrid method applicable to both tasks (Section \ref{sec:22}).

\begin{figure}[t]
\centering
\includegraphics[scale=0.8]{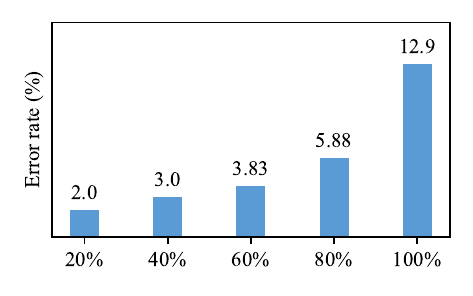}
\caption{Pseudo label error statistics. We report pseudo label error rates on 1,000 random images from CIFAR-10, using a 28-layer WideResNet model trained with 4,000 labeled samples only. The overall error rate is $12.90\%$, error rates for top-$20\%$, top-$40\%$, top-$60\%$, top-$80\%$ images according to prediction confidence are also plotted. It can be observed that high-confidence errors do exist and lots of data will be discarded to achieve a low error rate, e.g. $<3\%$ means discarding $60\%$ data. }
\label{fig2}
\end{figure}

Nevertheless, bootstrapping methods face a common issue, that is, pseudo supervisions tend to have classification errors. To address this, previous self-training methods \cite{lee2013pseudo,hung,cbst} select pseudo labels by confidence, e.g. the predicted probability from a model trained on the labeled subset. However, methods based on the common assumption that higher confidence corresponds to cleaner labels still have drawbacks. We conduct a pseudo labeling experiment (Fig. \ref{fig2}) to illustrate this issue.
As shown in Fig. \ref{fig2}, using confidence to select pseudo labels suffers from two limitations. First, low-confidence correct pseudo labels are often ignored, i.e. in order to achieve a low label error rate for pseudo supervision, a large portion of low-confidence correct pseudo labels have to be discarded. Second, high confidence errors do exist. We can observe that even pseudo labels with top-$20\%$ confidence still have some errors.
Moreover, pseudo supervision error from the model itself (termed as \textit{self-error}) can be extremely harmful in semi-supervised learning (Section \ref{sec:3}).

To address these limitations, we propose a novel method from  perspective of the inter-model disagreement. In particular, no matter what pseudo label selection metrics are employed, it is difficult for one model to find its own errors. Instead, two different models with disagreements on classification decisions, could potentially identify each other's errors. For instance, in image classification, model $A$ could provide a pseudo label on unlabeled image $x$ for model $B$ to learn, and we can quantify the disagreement between $A$ and $B$ by their prediction statistics (i.e. assigning lower loss/gradient to this image if their disagreement is larger). Since the possibility of different models confidently making the same mistakes is low, most incorrect pseudo labels will have limited impact on learning.

\begin{figure}[t]
\centering
\includegraphics[scale=0.28]{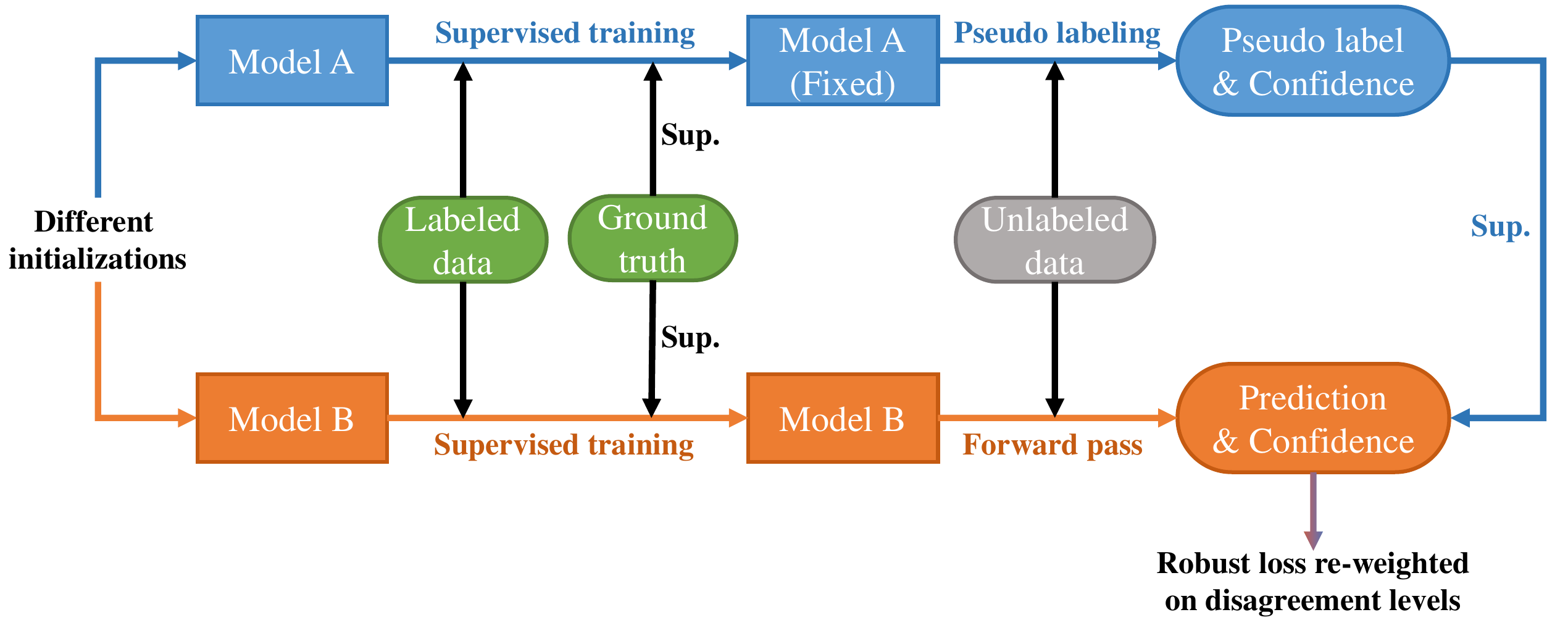}
\caption{Overview of our training framework. There are two different models trained on the labeled subset, and one model provides pseudo supervisions for the other. A noise-robust loss (dynamic loss) is introduced for learning on the unlabeled data, leveraging the inter-model disagreement based on two models' predictions and confidence. }
\label{fig1}
\end{figure}

Specifically, we propose Dynamic Mutual Training (DMT) with a noise-robust loss (Fig. \ref{fig1}). First, we instantiate two different models. Then, one model provides unreliable pseudo labels for the other on unlabeled data. In mutual training, We define three disagreement cases and corresponding loss re-weighting strategies, based on the relation of prediction confidence between the two models. In this way, loss is dynamically weighed to be lower when the disagreement is larger. Furthermore, inspired by 
curriculum learning, or \textit{easy to hard} \cite{bengio2009curriculum}, we apply DMT iteratively, each time considering more unlabeled data, to gradually increase performance.

Our main contributions are summarized as follows.
\begin{itemize}
    \item We analyze the pseudo label noise problem for semi-supervised learning and propose a new method from the new viewpoint of inter-model disagreement, i.e. instead of single model confidence, disagreement between models may indicate possible pseudo label errors.
    \item To quantify the inter-model disagreement, we propose a general and efficient bootstrapping approach, called Dynamic Mutual Training (DMT). DMT exploits the relation between different model predictions by a noise-robust loss function where a larger inter-model disagreement corresponds to a lower loss weighting. The performance of DMT is further enhanced by casting it into an iterative framework.
    \item We demonstrate the effectiveness of our approach in different tasks and datasets, i.e. semi-supervised image classification on CIFAR-10 and semi-supervised semantic segmentation on PASCAL VOC 2012 and Cityscapes. Through extensive comparisons and ablations, the proposed method shows state-of-the-art performance on both tasks. In the harder semantic segmentation task, our method even surpasses manual annotation under a certain setting.
\end{itemize}

\section{Related Work}
\label{sec:6}
\subsection{Semi-Supervised Learning}
\label{sec:61}
\begin{figure}[t]
\centering
\includegraphics[scale=0.8]{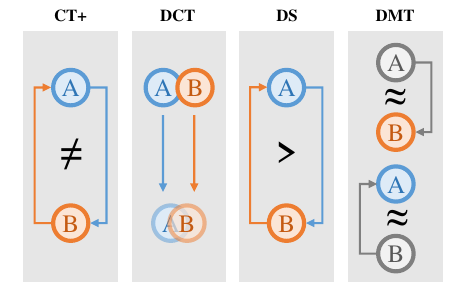}
\caption{Comparison of inter-model disagreement usages among Co-Teaching+ (CT+) \cite{yu2019does}, Deep Co-Training (DCT) \cite{qiao2018deep}, Dual Student (DS) \cite{ke2019dual} and DMT. \textbf{Left}: CT+ let models decide training examples for each other, but it only updates when models disagree ($\neq$). \textbf{Middle-Left}: DCT maximizes the agreement between two models (A, B). \textbf{Middle-Right}: DS let one model teach another when it is more certain on an example ($>$). \textbf{Right}: DMT updates one model with fixed (grey) pseudo labels from another, depending on how much they disagree ($\approx$). }
\label{fig7}
\end{figure}

 Here we focus on methods that are most relevant to DMT.  (Mainstream semi-supervised learning approaches including  entropy minimization \cite{lee2013pseudo,cascante2020curriculum,hung}, consistency regularization \cite{tarvainen2017mean,french2019semisupervised,ouali2020semi} and disagreement-based methods \cite{qiao2018deep,peng2020deep} will be formally described in the following Section \ref{sec:2}.) Among typical deep learning methods that use more than one models to explicitly/implicitly exploit the inter-model disagreement, the most similar to DMT are Dual Student \cite{ke2019dual} and Deep Co-training \cite{qiao2018deep}. In Dual Student, two models are trained in parallel online to select stable examples for each other to learn. However, the stable examples are decided by one model alone and the other model cannot dispute that supervision. In Deep Co-training, two models are also trained in parallel online, and the learning objective is to minimize their disagreement on the unlabeled subset. Since disagreement is minimized, the models can rapidly converge to the same set of weights online, thus explicit weight distance constrains have to be employed to avoid collapse. Largely different from them, the dynamic loss in DMT is determined by both models and does not require special constrains to avoid collapse. We illustrate the major differences between DMT and these two methods in Fig. \ref{fig7}.

Note that co-training methods \cite{qiao2018deep,peng2020deep} use different models and learn by maximizing their agreement on unlabeled data. In contrast, we view disagreement \cite{zhou2010semi} as a principle in this work, i.e. the inter-model disagreement provides \textbf{the possibility of learning}, i.e. the combined correct predictions are more than any of the two models. %
The exact method formulation of using it can vary. In this work, we employ the inter-model disagreement to specifically combat pseudo label noise by loss re-weighting, rather than penalizing disagreement in the learning objective.

Other than methodological differences to previous methods, DMT is also generally applicable to both image classification and semantic segmentation tasks. While most semi-supervised learning methods are ad-hoc and only work well in a limited range of tasks (e.g. only work on either image classification \cite{tarvainen2017mean} or pixel-wise task \cite{ke2020gct}). To the best of our knowledge, no previous methods have been shown to reach state-of-the-art performance on both image classification and semantic segmentation without additional efforts. For instance, consistency-based methods such as Mean Teacher \cite{tarvainen2017mean} works well in image classification but performs only comparable to Baseline on PASCAL VOC 2012 semantic segmentation (Tab. \ref{table:pascal}). High-dimensional perturbations have to be imposed to work in semantic segmentation, e.g. CutMix \cite{french2019semisupervised} and CCT \cite{ouali2020semi}.

\subsection{Learning with Noisy Labels}
\label{sec:62}
Learning with noisy labels is a well-studied topic. Most researches on this topic tackle random noise that can be modeled by a noise transition matrix, with each matrix entry as the label random flip probability from one class to another \cite{angluin1988learning,goldberger2017training}. We focus on general methods that have no explicit noise source modeling. Decoupling \cite{malach2017decoupling} trains two models simultaneously online, and only performs gradient descent when two models disagree, to decouple ``when'' and ``how'' to update model parameters, i.e. not allowing the noisy labels to control when to learn. Co-teaching \cite{han2018co} also trains two models online, while each model selects low loss examples for the other to train, and a similar policy has been exploited in semi-supervised learning by Dual Student \cite{ke2019dual}. Recently, Co-teaching+ \cite{yu2019does} combines the idea of Decoupling and Co-teaching. However, \textit{these methods still mostly show good performance regarding only random noise}.

Contrary to how disagreement is leveraged in Decoupling/Co-teaching+ \cite{malach2017decoupling,yu2019does}, in semi-supervised learning we deal with pseudo label noise made by deep nets similar to the model in training, where the disagreement between models signifies possible errors instead of an parameter update opportunity. Besides, Co-teaching \cite{han2018co} uses two models to select examples for each other, and the training targets are still determined by one model only. While in DMT the two models collaborate explicitly to re-weigh loss. Major differences between Co-teaching+ and DMT are illustrated in Fig. \ref{fig7}.

There is one approach that deals particularly with pseudo labels \cite{li2020density}. Specifically, given a feature embedding graph with every data point (image) as a node, a node's pseudo label can thus be rectified by its neighbors. It is feasible for relatively small datasets such as CIFAR-10, but rather unrealistic for segmentation datasets, where one image has millions of data points (pixels). Besides, the graph-based method is applied at the pseudo-labeling phase, making it complementary to DMT, which is adopted at training phase.

\section{Preliminaries}
\label{sec:2}

Semi-supervised learning methods are often based on certain prior knowledge, thus models can ``bootstrap'' themselves with extra unlabeled data for better generalization. There are mainly three types of prior knowledge.

\begin{enumerate}
    \item \textbf{Entropy minimization.} Predictive entropy is minimized for a model to make decisions on unlabeled data, which is apparently better than not making any decisions at all \cite{grandvalet2005semi}.
    \item \textbf{Consistency regularization.} Prediction should remain consistent when unlabeled data is perturbed by data augmentation \cite{tarvainen2017mean}.
    \item \textbf{Disagreement-based.} Multiple classifiers should reach an agreement on unlabeled data predictions \cite{qiao2018deep}.
\end{enumerate}

First, we focus on entropy minimization \cite{grandvalet2005semi} and summarize self-training methods as two types, online and offline. Then, we briefly describe the other two types of approaches: consistency regularization and disagreement-based methods. After that, we explain how consistency regularization can be integrated. Finally, we introduce the pseudo label noise problem, and how our method incorporating the inter-model disagreement can address it.

\subsection{Entropy Minimization and Self-training}
\label{sec:21}

Entropy $H$ is defined as:
\begin{align}
\label{eq:1}
    H = -\sum_{c=1}^{C} p_{c} \log p_{c}~,
\end{align}
where $p_{c}$ is the predicted probability for class $c$, and $C$ is the total number of classes. Since $\sum_{c=1}^{C} p_{c} = 1$, $H$ approaches the minimum value of $0$ when one class is $1$ and other classes are $0$. Thus, entropy minimization encourages the model to make a certain decision.

Self-training takes the most probable class as a pseudo label and train models on unlabeled data, which is a common approach to achieve the minimum entropy. \textit{Note that here we do not consider soft pseudo labels (labels as probability vectors instead of a hard label or one-hot vector), since they do not directly correspond to entropy minimization and we do not observe decent performance of using soft pseudo labels. } 

Denote $c^{\star} \leftarrow \argmax_c F(c|x)$, pseudo label $l$ is defined as:
\begin{align}
l& = 
    \begin{cases}
        c^{\star}, &F(c^{\star}|x) > T \\
        ignored, &otherwise~.
    \end{cases}
\end{align}
where $x$ is an image and $F(\cdot)$ is a classification model that predicts a probability distribution. $T$ is the threshold for selection (e.g. fixed value such as $0.5$ or ranked value such as the $20$-th percentile). We summarize self-training for semi-supervised learning as two types by \textit{when} pseudo labels are generated.

\textbf{Online self-training} generates pseudo labels after each forward pass in a network. Pseudo labels are directly selected based on some selection metric online and provide supervision for the immediate backward pass \cite{lee2013pseudo}.

\textbf{Offline self-training} firstly generates all pseudo labels, with a model trained only on the labeled subset (often followed by some form of selection process). Then the model is fine-tuned/re-trained on all labels (pseudo and human-labeled). This procedure can be applied iteratively by relabeling unlabeled data with the most recently trained model \cite{cbst}.

\begin{figure}[t]
\centering
\includegraphics[scale=1]{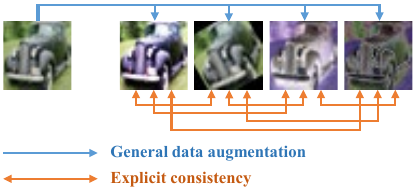}
\caption{Comparison between general data augmentation and explicit consistency regularization. It can be observed that they have similar principles and similar outcomes. General data augmentation can be seen as a simple form of ``anchoring''. Since the pseudo label on the unperturbed image has higher accuracy, it should be used as a semantic anchor for other perturbations. A similar concept of augmentation anchoring is also mentioned in \cite{berthelot2020remixmatch}. }
\label{fig8}
\end{figure}

\subsection{Consistency Regularization}

Consistency regularization perturbs unlabeled data randomly, so as to derive the consistency loss $\mathcal{L}_{con}$, usually in the form of mean squared error \cite{laine2017temporal}:

\begin{align}
\label{eq:r2}
    \mathcal{L}_{con} = ||F(x + \zeta) - F(x + \zeta')||_{2}^{2}~,
\end{align}
where $F$ is the model, $x$ is the input image. $\zeta$, $\zeta'$ are different perturbations, often achieved by random data augmentation. The recently popular variant Mean Teacher \cite{tarvainen2017mean} uses a teacher model $T$ whose weights are defined as the exponential moving average of $F$. %
$\mathcal{L}_{con}$ then becomes:

\begin{align}
\label{eq:r3}
    \mathcal{L}_{con} = ||F(x + \zeta) - T(x + \zeta')||_{2}^{2}~,
\end{align}

\subsection{Disagreement-based Methods}

Disagreement-based semi-supervised learning is formulated as Co-training \cite{qiao2018deep}, where prediction agreement is enforced between models. The loss for unlabeled data is Jensen-Shannon divergence (a common similarity metric):

\begin{align}
\label{eq:r4}
    \mathcal{L}_{cot} = H(\frac{1}{2}(p_{1} + p_{2})) - \frac{1}{2}(H(p_{1}) + H(p_{2}))~,
\end{align}
where $p_{1}$ and $p_{2}$ are predictions from two models. $H$ is entropy defined in Eq. \ref{eq:1}. To prevent the two models from collapsing to the same results, a diversity loss should be added, inspired by exploiting adversarial samples  \cite{peng2020deep}.

\subsection{Self-training and Consistency Regularization}
\label{sec:22}
We find that the consistency regularization plays a similar role to data augmentation in general. As shown in Fig. \ref{fig8}, with the same set of augmentation transforms, general data augmentation in offline self-training is similar to an ``anchored'' version of explicit consistency-based methods such as Mean Teacher \cite{tarvainen2017mean}. Anchoring may bring better performance overall if the anchor is positive, and degradation if the anchor is negative. In summary, consistency regularization and data augmentation have similar principles, and we observe good performance with data augmentation in offline self-training, which is the same as fully-supervised training after pseudo labeling.

However, for online self-training, it can be non-trivial to impose consistency regularization. Multiple different perturbations on input require multiple forward passes and induce higher computational cost, as demonstrated in many consistency-based methods \cite{laine2017temporal,tarvainen2017mean}. Online self-training is also more sensitive to pseudo label noise, which will be elaborated in Section \ref{sec:3}. Thus, we investigate the offline self-training case in our method. %

\subsection{Noisy Pseudo Label}
\label{sec:3}

Pseudo labels are noisy, since they are not generated by human annotators. Pseudo label noise is unique, coming from the model itself, different from random noise and noise from crowd-sourcing or search engines, which are widely explored and modeled according to the types and levels of noise \cite{angluin1988learning,goldberger2017training,malach2017decoupling}. Note that in other semi-supervised learning methods without explicit pseudo labels, self-supervision noise also exists. For instance, in consistency-based Mean Teacher \cite{tarvainen2017mean} where the Mean Squared Error (MSE) loss is used to enforce consistency of student model output to teacher model output, the pseudo-supervision (probability distribution) provided by the teacher is noisy. In this work, we discuss noisy pseudo labels (one-hot vectors) in self-training, which is more straightforward.

To address pseudo label noise, self-training methods adopt prediction confidence as noise indicator, i.e. selecting high confidence pseudo labels. The simplest policy is confidence thresholding, where only pseudo labels with confidence higher than a fixed threshold are selected \cite{lee2013pseudo}. Zou et al. further use different thresholds for each class \cite{cbst}. Hung et al. consider the confidence of a specialized two-class discriminator to discriminate between real and fake (wrong) labels \cite{hung}. Although noise rate is roughly lower when confidence is higher, it is difficult to attain sufficiently clean pseudo labels without discarding a large portion of unlabeled data (Fig. \ref{fig2}), proved by observations on semantic segmentation tasks \cite{hung} where only $27\% \sim 36\%$ pixels can be pseudo-labeled without performance degradation on PASCAL VOC 2012 and \cite{crst} where they only use $30\% \sim 50\%$ pseudo labels on Cityscapes. This goes against the purpose of semi-supervised learning, which is using rather than discarding unlabeled data. By contrast, we define different weights to samples instead of discarding them.

Other than wasting unlabeled data, selected pseudo labels still have some noise. This is particularly troublesome for online self-training. Because modern networks are trained with random and heavy data augmentation and the model in training changes after each parameter update, its predictions and errors also change. The change of errors can be extremely confusing when a model keeps minimizing entropy (fitting) on new errors, resulting in self-error accumulation throughout training. For offline self-training, although self-errors remain an obstacle, it is not as severe as that in online self-training, since the pseudo label errors are fixed throughout training.

However, there is always the same dilemma: a model can hardly correct itself. We introduce a new viewpoint from the inter-model disagreement, since it is possible to use one model to correct another. We assume that a same mistake made by different models simultaneously is not very likely, and the chances of different models all predicting the wrong class with high confidence are even less likely. In our proposed method, we quantify the disagreement between different models' predictions, allowing models to supervise each other with a disagreement re-weighted loss function. The proposed loss emphasizes on learning high-confidence agreements, effectively circumventing this dilemma.

\section{Method}
\label{sec:4}

In this section, we first present Dynamic Mutual Training (DMT) in Section \ref{sec:41} with the dynamic loss (Section \ref{sec:411}) and techniques for how to initialize models with disagreement in different tasks (Section \ref{sec:412}). Then we cast DMT to an iterative learning framework for better performance (Section \ref{sec:42}).

\begin{figure}[t]
\centering
\includegraphics[scale=0.20]{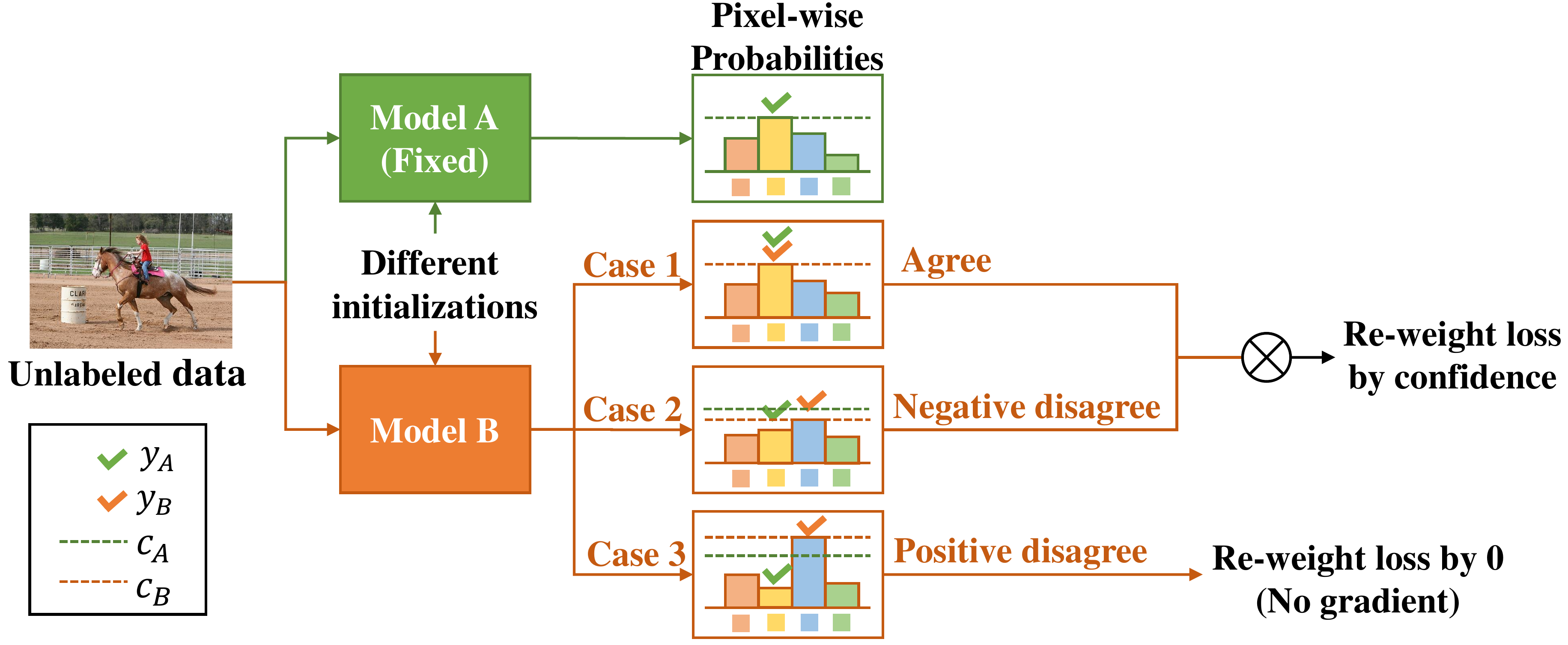}
\caption{Dynamic Mutual Training (DMT). There are two  different models ($F_{A}$, $F_{B}$). $F_{A}$ is used to generate pseudo labels ($y_{A}$) and record corresponding confidence ($c_{A}$) before mutual training (procedure marked by green lines), and $F_{B}$ (with prediction $y_{B}$ and confidence $c_{B}$) is then trained by them. There are three possible cases in mutual training and three corresponding loss re-weighting strategies based on the two models' disagreement degree, defined in Eq. \ref{eq:2}. All histograms above are for illustration purposes, \textit{rather than real outputs}. }
\label{fig4}
\end{figure}

\subsection{Dynamic Mutual Training}
\label{sec:41}

We propose Dynamic Mutual Training (DMT), to quantify the inter-model disagreement and enable noise-robust training, illustrated in Fig. \ref{fig4}. First, we train two different models $F_{A}$ and $F_{B}$ on the labeled subset from two different initializations/sub-samplings. Then, one model, e.g. $F_{A}$, is fixed and generates pseudo labels and confidence on the unlabeled subset. And the other, $F_{B}$, fine-tunes on all data (labeled and pseudo labeled) with our dynamically weighted cross-entropy loss. In the same way, $F_{B}$ can train $F_{A}$.

\subsubsection{The Dynamic Loss}
\label{sec:411}
We propose the dynamic loss, where the quantified inter-model disagreement serves as the dynamic loss weight. Taking image classification for example, we assume $F_{A}$ trains $F_{B}$, let $\mathcal{X}$, $\mathcal{U}$ denote labeled and unlabeled (pseudo labeled) samples in a batch of size $N$ and let $u$ be an unlabeled image in $\mathcal{U}$, we define its pseudo label as $y_{A} \leftarrow \argmax_y F_A(y|u)$, with confidence $c_{A} \leftarrow F_{A}(y_{A}|u)$. And the prediction in training is $y_{B} \leftarrow \argmax_y F_B(y|u)$, with confidence $c_{B} \leftarrow F_{B}(y_{B}|u)$. Let $p_{B} \leftarrow F_{B}(y_{A}|u)$ be the predicted probability of class $y_{A}$ by $F_{B}$. The dynamic loss weight $\omega_{u}$ is defined as:

\begin{align}
\label{eq:2}
    \omega_{u} =
    \begin{cases}
        p_{B}^{\gamma_{1}}, &y_{A} = y_{B} \\
        p_{B}^{\gamma_{2}}, &y_{A} \neq y_{B}, c_{A} \geq c_{B} \\
        0, &y_{A} \neq y_{B}, c_{A} < c_{B}.
    \end{cases}
\end{align}
The dynamic loss on unlabeled samples $\mathcal{L_{U}}$ is then defined as:
\begin{align}
\label{eq:3}
    \mathcal{L_{\mathcal{U}}} &= \frac{1}{N} \sum_{u, y_{A} \in \mathcal{U}} \omega_{u} CE\big(y_{A}, F_{B}(u)\big),
\end{align}
$CE(\cdot)$ is the cross-entropy loss:
\begin{align}
\label{eq:r8}
    CE(y, p) = -\sum_{c=1}^{C} y_{c} \log p_{c}~,
\end{align}
where $C$ is the total number of classes, $y$, $p$ are one-hot label vector and probability vector, respectively.

Intuitively, for the pseudo labeled data, there are three different cases in training: 
\begin{enumerate}
    \item \textbf{Agreement.} $F_{B}$ agrees with the pseudo label.
    \item \textbf{Negative disagreement.} $F_{B}$ disagrees with the pseudo label but the confidence on $F_{B}$'s decision is lower than the pseudo label's.
    \item \textbf{Positive disagreement.} $F_{B}$ disagrees with the pseudo label and has higher confidence.
\end{enumerate}
In cases 1 and 2, we use the current model's predicted probability $p_{B}$ on the pseudo labeled class as weight, perceived as the quantified disagreement, i.e. a higher $p_{B}$ means $F_{B}$ has a higher agreement with $F_{A}$. In case 3, we set the dynamic weight to $0$ because the pseudo label is probably incorrect.

Dynamic weights are further re-scaled by hyper-parameters $\gamma_{1},\gamma_{2}$, and a higher $\gamma$ magnifies confidence differences and suppresses gradients overall. For instance, assuming dynamic weights 0.9 and 0.5, the loss scale ratio is $0.9 / 0.5 = 1.8$; with $\gamma=2$, the new ratio becomes larger: $0.9^{2} / 0.5^{2} = 3.24$. It can be interpreted that a relatively larger $\gamma_{1}$ represents a more emphasized entropy minimization, a larger $\gamma_{2}$ represents a more emphasized mutual learning. Large $\gamma$ values are often better for high-noise scenarios, or to maintain larger inter-model disagreement.

Note that training uses the labeled subset along with the pseudo-labeled data, and the loss for labeled data $\mathcal{L}_{\mathcal{X}}$ remains unchanged, i.e. the typical cross-entropy loss:
\begin{align}
\label{eq:4}
    \mathcal{L_{\mathcal{X}}} = \frac{1}{N} \sum_{x, gt \in \mathcal{X}} CE\big(gt, F_{B}(x)\big),
\end{align}
where $x$ and $gt$ denote image and ground truth pairs. The combined loss $\mathcal{L}$ is defined as:
\begin{align}
\label{eq:5}
    \mathcal{L} = \mathcal{L_{\mathcal{X}}} + \mathcal{L_{\mathcal{U}}}.
\end{align}

The above example is given on classification. With regard to semantic segmentation, $\omega_{u}^{H \times W}$ is a pixel-wise map ($H$ for height and $W$ for width), the re-weighting strategy remains the same and applies on each pixel.

\subsubsection{Initialize Disagreement}
\label{sec:412}

A key problem for leveraging the inter-model disagreement is how to initialize sufficiently different models. For simpler tasks such as CIFAR-10 image classification, it is simple to just randomly initialize different models for sufficient disagreement. However, for tasks that require pre-trained weights to work well, e.g. semantic segmentation, sufficiently different off-the-shelf pre-trained weights are hard to obtain, and the extra amount of time needed for a new pre-training is too costly compared to the task at hand. Thus, we mainly use pre-trained weights from different datasets (ImageNet and COCO) in semantic segmentation. For some extreme cases where labeled data is scarce and one set of pre-trained weights is clearly superior (about 100-200 labeled images), we use the better pre-trained weights, and train two models from different sub-subsets of the labeled subset by the following method.

\noindent \textbf{Difference maximized sampling.} Let $\mathcal{R}$ be the randomly shuffled labeled subset with size $L$. We aim to sample two equal-sized sub-subsets $\mathcal{S}_{A}$, $\mathcal{S}_{B}$ from $\mathcal{R}$, each with size $\alpha L~(0.5 < \alpha < 1)$. The goal (maximized difference) is to have the smallest intersection for them. We can simply derive $\mathcal{S}_{A}$, $\mathcal{S}_{B}$ as:
\begin{align}
\label{eq:r11}
\begin{cases}
    \mathcal{S}_{A} &= \mathcal{R}_{0: \alpha L}\\
    \mathcal{S}_{B} &= \mathcal{R}_{(1 - \alpha) L : L}
\end{cases}
\end{align}
where $\mathcal{R}_{s:e}$ represents a selection by index from $s$ to $e$, including $s$.

\subsection{Iterative Framework}
\label{sec:42}

In this section we cast DMT into an iterative framework for better performance. Curriculum learning \cite{bengio2009curriculum}, or \textit{easy to hard}, has been explored in semi-supervised image classification \cite{cascante2020curriculum} and unsupervised domain adaptive semantic segmentation (a similar setup to semi-supervised learning) \cite{cbst} for better bootstrapping performance. Specifically, the same bootstrapping algorithm is repeated for multiple iterations, each iteration explores a harder setting, e.g. more pseudo labels with lower confidence. Inspired by their successes, we also perform DMT iteratively to achieve better performance.

\BlankLine

\scalebox{0.7}{
\begin{minipage}{1.35\linewidth}
\begin{algorithm}[H]
\DontPrintSemicolon
  \KwIn{Unlabeled subset $\mathcal{S}_{u}$, labeled subset $\mathcal{S}_{l}$.}
  \KwOut{Final best model $F$.}

  \BlankLine
  \caption{Pseudo code for iterative DMT process in image classification.}\label{alg:1}
    
  Randomly initialize $F^{0}$\;
  Train $F^{0}$ on $\mathcal{S}_{l}$\;
  
  $\alpha = \{20\%, 40\%, 60\%, 80\%, 100\%\}$\;
  
  \ForEach{iteration $i \in \{1, 2, 3, 4, 5\}$}{
    Pseudo labeled set $\mathcal{S}_{p}$ $\leftarrow$ Predict and save top $ \alpha_{i}$ images on $\mathcal{S}_{u}$ with $F^{i-1}$\;
    Randomly initialize $F^{i}$ with a previously unused random seed\;
    Train $F^{i}$ on both $\mathcal{S}_{l}$ and latest $\mathcal{S}_{p}$ with the dynamic loss\;
  }
  $F$ = $F^{5}$\;
  
\end{algorithm}
\end{minipage}
}

\subsubsection{Image Classification}
\label{sec:421}

For this task, we first train on the labeled subset, then conduct DMT iteratively for multiple times; each time we select more top-confident pseudo labels from the unlabeled subset and re-train a randomly initialized model for sufficient disagreement, same as concurrent work Curriculum Labeling \cite{cascante2020curriculum}. Pseudo code is shown in Alg. \ref{alg:1}. However, the model re-trained from scratch provides little meaningful information at early training stage, thus we use a sigmoid-like function for $\gamma$ values inspired by \cite{tarvainen2017mean}. Concretely, with the total training steps $t_{max}$, at step $t$, $\gamma = \gamma_{max} e^{5 (1-\frac{t}{t_{max}})^2}$.

\BlankLine

\scalebox{0.7}{
\begin{minipage}{1.35\linewidth}
\begin{algorithm}[H]
\DontPrintSemicolon
  \KwIn{Unlabeled subset $\mathcal{S}_{u}$, labeled sub $\mathcal{S}_{l}$.}
  \KwOut{Final best model $F$.}

  \BlankLine
  \caption{Pseudo code for iterative DMT process in semantic segmentation.}\label{alg:2}
  
  \uIf{start from different pre-trained weights}{
    Initialize $F_{A}^{0}$ and $F_{B}^{0}$ with different pre-trained weights\;
    Train $F_{A}^{0}$ on $\mathcal{S}_{l}$\;
    Train $F_{B}^{0}$ on $\mathcal{S}_{l}$\;
  }
  \Else{
    Initialize $F_{A}^{0}$ and $F_{B}^{0}$ with the same pre-trained weights\;
    $\mathcal{R} \leftarrow$ Randomly shuffled $\mathcal{S}_{l}$\;
    $\mathcal{S}_{A}$, $\mathcal{S}_{B}$ = DifferenceMaximizedSampling($\mathcal{R}$)  // see Eq. \ref{eq:r11}\;
    Train $F_{A}^{0}$ on $\mathcal{S}_{A}$\;
    Train $F_{B}^{0}$ on $\mathcal{S}_{B}$\;
  }

  $\alpha = \{20\%, 40\%, 60\%, 80\%, 100\%\}$\;
  
  \ForEach{iteration $i \in \{1, 2, 3, 4, 5\}$}{
    Pseudo labeled set $\mathcal{S}_{p}$ $\leftarrow$ Predict and save top $ \alpha_{i}$ pixels from each classes on $\mathcal{S}_{u}$ with $F_{A}^{i-1}$\;
    Fine-tune $F_{B}^{i}$ from $F_{B}^{i-1}$ on both $\mathcal{S}_{l}$ and latest $\mathcal{S}_{p}$ with the dynamic loss\;
    Pseudo labeled set $\mathcal{S}_{p}$ $\leftarrow$ Predict and save top $ \alpha_{i}$ pixels from each classes on $\mathcal{S}_{u}$ with $F_{B}^{i-1}$\;
    Fine-tune $F_{A}^{i}$ from $F_{A}^{i-1}$ on both $\mathcal{S}_{l}$ and latest $\mathcal{S}_{p}$ with the dynamic loss\;
  }
  
  $F$ = The best model from $F_{A}^{5}$, $F_{B}^{5}$, in term of mean IoU\;
  
\end{algorithm}
\end{minipage}
}

\subsubsection{Semantic Segmentation}
\label{sec:422}

\begin{figure}[t]
\centering
\includegraphics[scale=0.7]{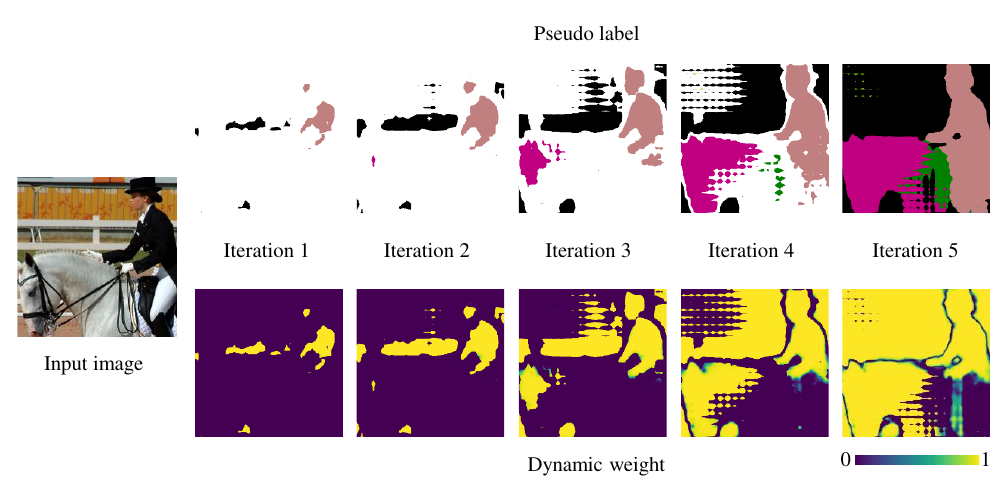}
\caption{Illustration of iterative DMT training on PASCAL VOC 2012. We show pseudo labels from one model and dynamic weights generated when training another. At each iteration, more pixels are pseudo labeled. For some apparently incorrect regions we can observe rather low dynamic weights. Models are trained with 1/20 manually labeled data in the training set. White regions are ignored in labels and corresponding dynamic weights are shown as 0. }
\label{fig12}
\end{figure}

Motivated by the fact that some classes are much easier to learn than others in semantic segmentation, CBST \cite{cbst} proposed an iterative self-training scheme by using more top-confident pseudo labels \textit{from each class} at each iteration. Furthermore, unlike image classification, fine-tuning performs reasonably well in semantic segmentation, and converges faster than re-training. Inspired by CBST, we conduct two separate fine-tunings between two differently initialized models, as shown in Alg. \ref{alg:2}. In this setup, two models train each other equally and the inter-model disagreement is more straightforwardly utilized. To better illustrate how DMT is used iteratively for this harder task, we provide some examples of pseudo labels and dynamic weights during training in Fig. \ref{fig12}.

However, there is some difference between our iterative framework and CBST, since CBST does not select top-confident pseudo labels by direct ranking like us. Instead, it first uses class-wise thresholds (defined by ranking) to re-normalize softmax predictions by dividing the thresholds class-wise, then pixels with confidence over 1 are selected. In most cases, this is the same as direct ranking, while in extreme cases, the predicted class would change, e.g. for a binary classification task, softmax result [0.6, 0.4] that originally predicts class 0, re-normalized by ranked thresholds [0.61, 0.39], will be changed to prediction class 1. This kind of mistake happen only when using nearly all pseudo labels, which does not affect the original CBST experiments since they use less than half pseudo labels. While we use up to all data, re-normalization brings a slight degradation to final performance. Therefore, we use direct ranking instead of re-normalization.

We set 5 iterations to our method for training. Since fine-tuning converges very fast, the total number of training steps for each one of the two models remains similar to a fully-supervised training on the entire dataset.

\section{Experiments}
\label{sec:5}

In this section, we first specify dataset configurations (Section \ref{sec:51}) and implementation details (Section \ref{sec:52}). Then we compare the proposed DMT with state-of-the-art methods on image classification and semantic segmentation (Section \ref{sec:53}). %
Finally, we analyze DMT by conducting a set of ablation studies (Section \ref{sec:54}).

\subsection{Datasets}
\label{sec:51}

For image classification, we employ the commonly used CIFAR-10 \cite{krizhevsky2009learning} dataset. For semantic segmentation, we evaluate our method on the popular PASCAL VOC 2012 \cite{everingham2015pascal} and Cityscapes \cite{cordts2016cityscapes} datasets.

\begin{itemize}
\item \textbf{CIFAR-10.} The CIFAR-10 \cite{krizhevsky2009learning} dataset has 10 classes, 50,000 training samples and 10,000 test samples. Most methods extract 5,000 samples from the training set to use as validation. While we consider using a realistic \cite{oliver2018realistic} validation set with 200 samples (see \ref{sec:a4} for how we validate our method properly with a limited validation set).

\item \textbf{PASCAL VOC 2012.} The original PASCAL VOC 2012 \cite{everingham2015pascal} semantic segmentation dataset has 21 classes, 1,464 training samples and 1,449 validation samples (\textit{val}) featuring common objects. We use the SBD \cite{hariharan2011semantic} augmented version with 10,582 training samples following common practice.

\item \textbf{Cityscapes.} The Cityscapes \cite{cordts2016cityscapes} dataset has 19 classes, 2,975 training samples, and 500 validation samples (\textit{val}) for urban driving scenes. Following \cite{mittal2019semi}, we down-sample the images by half to $512 \times 1024$.
\end{itemize}

\subsection{Implementation Details}
\label{sec:52}
Since DMT in an iterative framework has better performance and overall fast convergence as described in Section \ref{sec:42}, all DMT experiments are conducted with the default 5 iterations. Our method is implemented on PyTorch with mixed-precision training \cite{DBLP:conf/iclr/MicikeviciusNAD18}. All experiments were conducted on a single RTX-2080 Ti GPU. The fully-supervised learning result on the entire dataset is termed as Oracle, i.e. the performance upper-bound for semi-supervised learning. However, this upper-bound only holds when human annotations have good quality across the entire dataset. We show how this supposed upper-bound can be surpassed by our semi-supervised learning in Section \ref{sec:532}.

\subsubsection{Network Architectures}
\label{sec:521}

\begin{itemize}
    \item \textbf{Image classification.} We follow MixMatch \cite{berthelot2020remixmatch} and use a shallow residual network WideResNet-28-2 (WRN-28-2) \cite{zagoruyko2016wide} as the backbone.
    \item \textbf{Semantic segmentation.} We follow \cite{mittal2019semi,hung} and use DeepLab-v2 ResNet-101 \cite{deeplabv2} as the backbone, without multi-scale fusion or CRF post-processing. Our implementation has slightly better performance than previous works, which is better aligned with the original DeepLab-v2 paper ($74.75\%$ averaged mean IoU on PASCAL VOC 2012 \textit{val} set, higher than $73.6\%$ reported in \cite{hung}).
\end{itemize}

\subsubsection{Training}
\label{sec:522}

\begin{itemize}
    \item \textbf{Image classification.} Each DMT iteration has 750 epochs with a learning rate of $0.1$, a weight decay of $5 \times 10^{-4}$, a momentum of $0.9$, the cosine annealing technique and a batch size of 512, which is the same as Curriculum Labeling \cite{cascante2020curriculum}; we do not use SWA \cite{izmailov2018averaging} for fair comparisons with other methods. Data augmentations are RandAugment \cite{cubuk2020randaugment} with Cutout. We randomly select one augmentation operation with random intensity at each step to avoid hyper-parameter tuning (number of operations $n$ and intensity $m$ are hyper-parameters in RandAugment). We also use mixup \cite{zhang2018mixup} by interpolating dynamic weights along with input images.
    \item \textbf{Semantic segmentation.} Each DMT iteration has less training steps due to fine-tuning. We use SGD with a momentum of $0.9$, the \emph{poly} learning rate schedule and a batch size of 8. Data augmentations include random scaling, random cropping and random flipping. We train and pseudo label at a spatial resolution of $321 \times 321$ (PASCAL VOC 2012) and $256 \times 512$ (Cityscapes).
\end{itemize}

To avoid too much hyper-parameter tuning, we set $\gamma_{1} = \gamma_{2}$. Pseudo-labeled data are used along with labeled data in DMT, thus we have a certain ratio (labeled : unlabeled, e.g. 1 : 7) to combine them in a batch. More details are listed in Tab. \ref{table:exp}.

\begin{table}
\caption{Hyper-parameter settings.}
\label{table:exp}
\centering
\resizebox{12cm}{!}{
\begin{tabular}{cccccccccccc} \toprule

& dataset & \makecell{labeled \\ ratio} & $\gamma_{1}$ & $\gamma_{2}$ & \makecell{learning \\ rate} & training & epochs & \makecell{batch \\ size} & \makecell{batch \\ ratio} & augmentations \\

\midrule

1 & PASCAL VOC & 1/8 & 5 & 5 & $1 \times 10^{-3}$ & fine-tuning & 5 & 8 & 7:1 & \multirow{6}{*}{\makecell{random scale \\ random crop \\ random horizontal flip}} \\

2 & PASCAL VOC & 1/20 & 5 & 5 & $1 \times 10^{-3}$ & fine-tuning & 4 & 8 & 7:1 & \\

3 & PASCAL VOC & 1/50 & 5 & 5 & $1 \times 10^{-3}$ & fine-tuning & 4 & 8 & 7:1 & \\

4 & PASCAL VOC & 1/106 & 5 & 5 & $1 \times 10^{-3}$ & fine-tuning & 4 & 8 & 7:1 & \\

5 & Cityscapes & 1/8 & 3 & 3 & $4 \times 10^{-3}$ & fine-tuning & 10 & 8 & 3:1 & \\

6 & Cityscapes & 1/30 & 3 & 3 & $4 \times 10^{-3}$ & fine-tuning & 8 & 8 & 7:1 & \\ \midrule

7 & CIFAR-10 & 4k labels & 4 & 4 & $1 \times 10^{-1}$ & re-training & 750 & 512 & 7:1 & \multirow{2}{*}{\makecell{random augmentation \\ with random intensity}} \\

8 & CIFAR-10 & 1k labels & 4 & 4 & $1 \times 10^{-1}$ & re-training & 750 & 512 & 31:1 & \\

\bottomrule
\end{tabular}
}
\end{table}

\subsubsection{Testing}
\label{sec:523}

\begin{itemize}
    \item \textbf{Image classification.} We report the 5-times averaged \textit{test} set performance with an exponential moving averaged (EMA) network on CIFAR-10 following MixMatch \cite{berthelot2019mixmatch}.
    \item \textbf{Semantic segmentation.} We report the three-times averaged \textit{val} set mean intersection-over-union (mean IoU) in semantic segmentation tasks following common practice, provided the \textit{test} set labels for these datasets are not publicly available.
\end{itemize}

\subsection{Comparisons}
\label{sec:53}
To show the effectiveness and generality of DMT, we compare it with state-of-the-art methods on both image classification and semantic segmentation benchmarks as detailed in Section \ref{sec:51}: CIFAR-10 \cite{krizhevsky2009learning}, PASCAL VOC 2012 \cite{everingham2015pascal} and Cityscapes \cite{cordts2016cityscapes}. Standard practice for evaluating semi-supervised learning on these datasets is to treat most of a dataset as the unlabeled subset and use a small portion as the labeled subset.

\subsubsection{CIFAR-10}
\label{sec:531}

For CIFAR-10, we compare our method with consistency-based Mean Teacher (MT) \cite{tarvainen2017mean}, self-training method Curriculum Labeling (CL) \cite{cascante2020curriculum}, methods explicitly/implicitly using multiple models, Deep Co-Training between two models (DCT) \cite{qiao2018deep} and Dual Student (DS) \cite{ke2019dual}, strong hybrid method MixMatch \cite{berthelot2019mixmatch}, and the combination of graph-based pseudo label propagation techniques in Density Aware Graph-based framework (DAG) \cite{li2020density}. Methods are evaluated on the commonly adopted 1,000 labels and 4,000 labels splits. Supervised performance with mixup and strong data augmentation on the labeled subset is reported as Baseline. Baseline, CL (our re-implementation, without SWA) and DMT are implemented in the same codebase, while for other methods with WRN-28-2 we take the reported numbers from MixMatch. The remaining numbers are taken from the original papers.

As shown in Tab. \ref{table:more}, DMT steadily improves CL with the dynamic loss, larger performance gain is shown in harder setting (smaller labeled subset). Note that CL is already a very high-performance method (only $2.33\%$ less than Oracle performance using 4,000 labels), it is rather difficult to gain further improvements in a strictly controlled comparison such as ours. While our method still achieves a slight improvement. We also compare DMT with more methods in Tab. \ref{table:cifar}, where DMT shows better performance than state-of-the-art methods. We are aware that recently ReMixMatch \cite{berthelot2020remixmatch} has obtained an accuracy of $94.86\%$, which is certainly benefited from using multiple forward passes and multiple losses, e.g. rotation loss, thus takes much more computing to train.

\begin{table}[t]
\caption{Results ($\%$) between DMT and CL on CIFAR-10 \textit{test} set. }
\label{table:more}
\centering
\scalebox{0.7}{
\begin{tabular}{cccc} \toprule
 & Baseline & CL \cite{cascante2020curriculum} & DMT \\
\midrule

4000 labels & 86.08 & 94.02 & 94.21 (\textbf{+0.19}) \\
1000 labels & 75.14 & 90.61 & 91.51 (\textbf{+0.90}) \\

\bottomrule
\end{tabular}
}
\end{table}

\begin{table}[t]
\caption{Results ($\%$) for DMT and other methods on CIFAR-10 \textit{test} set using 4,000 labels. Oracle performance is $96.35\%$. }
\label{table:cifar}
\centering
\resizebox{12cm}{!}{
\begin{tabular}{c|c|c|c|c|c|c|c|c} \toprule
method & Baseline & MT \cite{tarvainen2017mean} & DCT \cite{qiao2018deep} & DS \cite{ke2019dual} & MixMatch \cite{berthelot2019mixmatch} & DAG \cite{li2020density} & CL \cite{cascante2020curriculum} & Ours (DMT) \\
\midrule
network & WRN-28-2 & WRN-28-2 & CNN-13 & CNN-13 & WRN-28-2 & CNN-13 & WRN-28-2 & WRN-28-2 \\ 
accuracy & 86.08 & 89.64 & 90.97 & 91.11 & 93.76 & 93.87 & 94.02 & \textbf{94.21} \\
\bottomrule
\end{tabular}
}
\end{table}

\subsubsection{PASCAL VOC 2012}
\label{sec:532}

PASCAL VOC 2012 is the most commonly used benchmark for semi-supervised semantic segmentation. We compare our method with consistency-based Mean Teacher directly adapted to semantic segmentation (MT-Seg), Mean Teacher with strong CutMix augmentation \cite{french2019semisupervised}, feature-level consistency-based method Cross-Consistency Training (CCT) \cite{ouali2020semi}, adapted Dual Student for semantic segmentation with an auxiliary flaw detector (GCT) \cite{ke2020gct}, GAN-based method \cite{hung} that pre-trains a discriminator to select pseudo labels, and hybrid method s4GAN + MLMT \cite{mittal2019semi} that adds consistency regularization upon \cite{hung} by an extra classification branch. Methods are evaluated on 4 challenging splits: 1/106 (100 labels), 1/50, 1/20 and 1/8. We do not use more than 1/8 data which is becoming easier and pose less challenge to state-of-the-art methods. Supervised performance on the labeled subset is reported as Baseline. MT-Seg, CCT and GCT performance are the re-evaluated results in the GCT codebase\footnote{https://github.com/ZHKKKe/PixelSSL/tree/master/task/sseg}; others are taken from the original papers. All methods use the same network architecture as ours except for CCT in which a slightly superior architecture PSPNet-ResNet-101 \cite{zhao2017pyramid} is used for evaluation.

\begin{table}[t]
\caption{Mean IoU ($\%$) results for DMT and other methods on PASCAL VOC 2012 \textit{val} set. Performance gap to Oracle is shown in brackets. \textit{$\dag$ Updated numbers from s4GAN + MLMT. * ImageNet pre-training.}}
\label{table:pascal}
\centering
\resizebox{12cm}{!}{
\begin{tabular}{c|c|ccccc} \toprule
method & network & 1/106 & 1/50 & 1/20 & 1/8 & Oracle \\
\midrule
Baseline & DeepLab-v2 & 46.66 (-28.09) & 55.62 (-19.13) & 62.29 (-12.46) & 67.37 (-7.38) & 74.75 \\
MT-Seg \cite{tarvainen2017mean} & DeepLab-v2 & - & - & - & 67.65 (-5.94) & 73.59 \\
Hung et al. \cite{hung} & DeepLab-v2 & - & 57.2$^{\dag}$ (-17.7) & 64.7$^{\dag}$ (-10.2) & 69.5 (-5.4) & 74.9 \\
s4GAN + MLMT \cite{mittal2019semi} & DeepLab-v2 & - & 63.3 (-12.3) & 67.2 (-8.4) & 71.4 (-4.2) & \textbf{75.6} \\
CutMix \cite{french2019semisupervised}* & DeepLab-v2 & 53.79 (-18.75) & 64.81 (-7.73) & 66.48 (-6.06) & 67.60 (-4.94) & 72.54 \\
CCT \cite{ouali2020semi} & PSPNet & - & - & -	& 70.45 (-4.80) & 75.25 \\
GCT \cite{ke2020gct} & DeepLab-v2 & - & - & -	& 70.57 (-3.49) & 74.06 \\
Ours (DMT) & DeepLab-v2 & \textbf{63.04 (-11.71)} & \textbf{67.15 (-7.60)} & \textbf{69.92 (-4.83)} & \textbf{72.70 (-2.05)} & 74.75 \\
\bottomrule
\end{tabular}
}
\end{table}

\begin{figure}[t]
\centering
\includegraphics[scale=0.8]{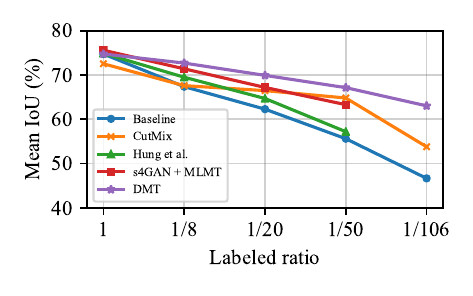}
\caption{Comparison of semi-supervised semantic segmentation performance on PASCAL VOC 2012 with multiple data splits. The proposed DMT has more stable performance across different labeled ratios. }
\label{fig5}
\end{figure}

As shown in Tab. \ref{table:pascal}, DMT outperforms other methods with a clear margin. However, some methods use GAN and extra network branch, or have implementation flaws, resulting in different Oracle and Baseline performances. Thus, we further show performance gaps to Oracle in brackets for fair comparisons, where DMT's performance is also the closest to Oracle. Our proposed DMT is the only method showing a clear improvement over Baseline on the challenging 100 labels split other than CutMix (1/106 in Tab. \ref{table:pascal}). In addition, DMT shows more stable performance across different labeled ratios (Fig. \ref{fig5}).

\textbf{Comparing with human supervision.} We design an interesting experiment to show how DMT is even superior to human annotators. Specifically, the original PASCAL VOC 2012 dataset only labels 1,464 training images, called the \textit{train} set. While the commonly used 10,582 training set \textit{trainaug} contains 9,118 images from SBD \cite{hariharan2011semantic}. The SBD dataset uses the same set of images as PASCAL VOC and annotates object outlines by Amazon Mechanical Turk (AMT), which can be filled as segmentation masks. However, unprofessional annotators from AMT tend to draw coarse outlines. Thus, \textit{trainaug} has worse label quality than \textit{train}. Therefore, we use \textit{train} as the labeled subset and the 9,118 images from SBD as the unlabeled subset (by removing the SBD labels), and experiment DMT with the same hyper-parameters in the 1/8 split experiment. Surprisingly, as shown in Tab. \ref{table:limit}, DMT exceeds the performance of Oracle (fully-supervised training on \textit{trainaug}). This suggests that DMT renders human supervision at this quality (AMT) unnecessary for semantic segmentation, except for somewhat faster training since DMT needs two models (roughly twice the training budget of one model).

\begin{table}[t]
\caption{Mean IoU ($\%$) comparisons between Oracle and DMT on PASCAL VOC 2012 1464/9118 split. \textit{val} mean IoU ($\%$) reported.}
\label{table:limit}
\centering
\scalebox{0.7}{
\begin{tabular}{c|ccc} \toprule
& number of images & number of labels & mean IoU \\
\midrule
Baseline & 1464 & 1464 & $72.10 \pm 0.53$ \\
DMT & 10582 & 1464 & $\mathbf{74.85 \pm 0.29}$ \\
Oracle & 10582 & 10582 & $74.75 \pm 0.25$ \\

\bottomrule
\end{tabular}
}
\end{table}

\begin{figure}[t]
\centering
\includegraphics[scale=0.6]{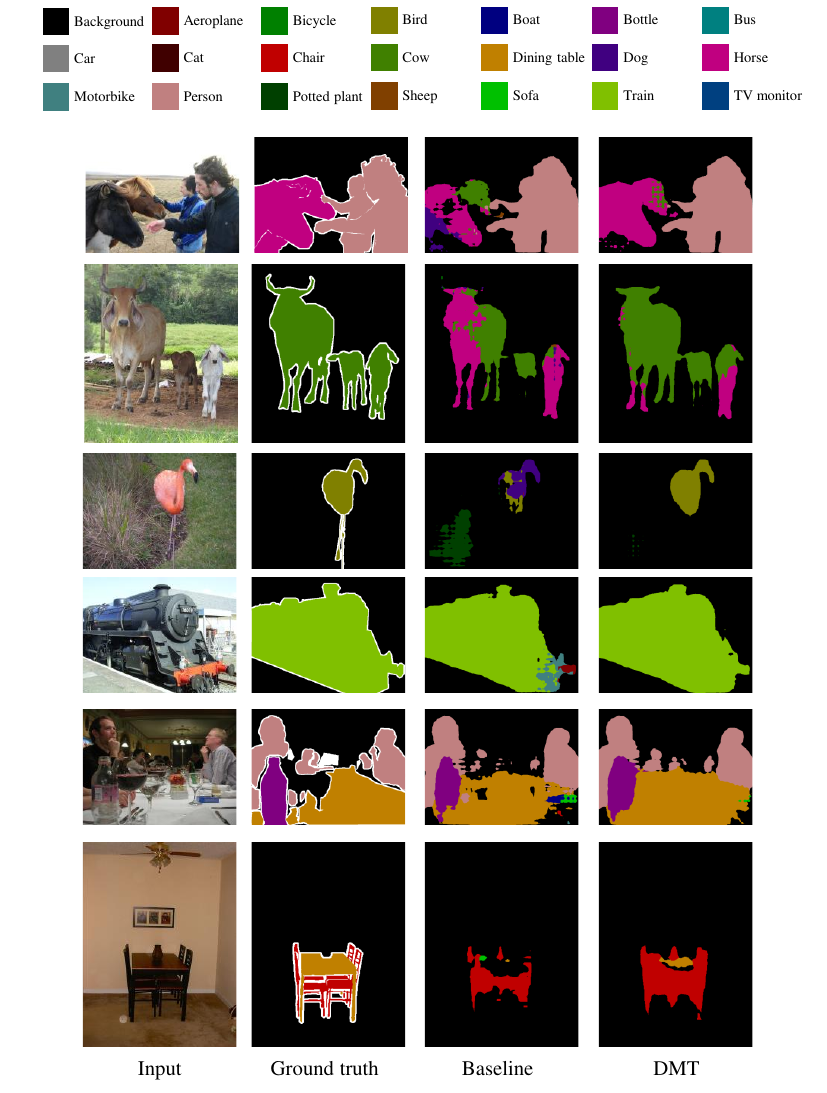}
\caption{Qualitative results on PASCAL VOC 2012. Models are trained with 1/20 pixel-level labeled data in the training set. White regions are ignored in ground truth. }
\label{fig10}
\end{figure}

\textbf{Qualitative results.} We provide qualitative comparisons in segmentation results among Baseline, DMT and ground truth. As shown in Fig. \ref{fig10}, there is confusion between similar classes in Baseline predictions (column 3), such as \textit{horse} and \textit{cow} (row 1, 2), \textit{dog} and \textit{bird} (row 3), \textit{train} and \textit{motorbike} (row 4). After dynamic mutual training (column 4), class confusion is mostly resolved. Also finer details are recovered (e.g. distant people in row 5 and chairs in row 6). Moreover, Baseline entirely fails to detect the \textit{dining table} in row 6.

\subsubsection{Cityscapes}
\label{sec:533}

Cityscapes features complex street scenes which are less ventured by semi-supervised learning methods. Of the six methods evaluated in Section \ref{sec:532}, three methods have chosen to report performance on this dataset. We follow the same evaluation protocols as PASCAL VOC 2012, to evaluate each method on 1/30 (100 labels) and 1/8 splits. All reported numbers are taken from the original papers.

As shown in Tab. \ref{table:city}, Hung et al. \cite{hung} is merely comparable to our fully-tuned Baselines. Although stronger methods s4GAN + MLMT \cite{mittal2019semi} and CutMix \cite{french2019semisupervised} still obtain good results, DMT outperforms them by $2 \sim 3\%$. Refer to \ref{sec:a3} for proper baseline training and \ref{sec:a1} for qualitative results.

\begin{table}[t]
\caption{Mean IoU ($\%$) results for DMT and other methods on Cityscapes \textit{val} set. Performance gap to Oracle is shown in brackets. \textit{* ImageNet pre-training.}}
\label{table:city}
\centering
\scalebox{0.7}{
\begin{tabular}{c|ccc} \toprule
method & 1/30 & 1/8 & Oracle \\
\midrule
Baseline & 49.54 (-18.62) & 59.65 (-8.51) & \textbf{68.16} \\
Hung et al. \cite{hung} & - & 58.8 (-8.9) & 67.7 \\
s4GAN + MLMT \cite{mittal2019semi}* & - & 59.3 (-6.5) & 65.8 \\
CutMix\cite{french2019semisupervised}* & 51.20 (-16.48) & 60.34 (-7.34) & 67.68 \\
Ours (DMT) & \textbf{54.80 (-13.36)} & \textbf{63.03 (-5.13)} & \textbf{68.16} \\
\bottomrule
\end{tabular}
}
\end{table}

\subsection{Ablations}
\label{sec:54}

To further validate our method choice and provide additional insights for online and offline self-training, we conduct the following ablations.

\begin{itemize}
    \item \textbf{Online ST.} Instead of 5 iterations of DMT, we perform online self-training with fixed confidence threshold 0.9 for 20 epochs.
    \item \textbf{CBST.} The CBST algorithm \cite{cbst} is modified to direct ranking, i.e. iterative class-balanced self-training without dynamic loss, similar to a class-balanced version of CL \cite{cascante2020curriculum}.
    \item \textbf{DST.} Same as DMT, except using only one model to provide pseudo labels for itself, i.e. DMT with only one model $F_{A}$ fine-tuning itself.
    \item \textbf{DMT-Naive.} We directly re-weight the loss by confidence without distinguishing the three cases in Eq. \ref{eq:2}.
    \item \textbf{DMT-Flip.} In the third condition of Eq. \ref{eq:2}, since the pseudo label is likely incorrect, instead of setting loss to 0, we flip the pseudo label to the current model's prediction and weight the loss by $(1 - c_{A})^{\gamma_{2}}$, acting as an estimate of disagreement between models, given the pseudo label is flipped.
\end{itemize}

\begin{table}[t]
\caption{Ablations on PASCAL VOC 2012 (one random 1/20 split and one random 1/50 split). \textit{val} mean IoU ($\%$) is reported. }
\label{table:ablations}
\centering
\scalebox{0.7}{
\begin{tabular}{c|ccccccc} \toprule
ablations & Baseline & Online ST & CBST & DST & DMT-Naive & DMT & DMT-Flip \\
\midrule
1/20 & 61.90 & 63.12 & 65.09 & 69.43 & 70.00 & 70.16 & 70.17 \\
1/50 & 56.29 & 53.52 & 62.29 & 66.50 & 64.95 & 68.37 & 68.35 \\
\bottomrule
\end{tabular}
}
\end{table}

To show clear differences between setups, the ablations are carried out on PASCAL VOC 2012 in Tab. \ref{table:ablations}. This dataset is sufficiently complex and experiments run faster due to fine-tuning.

\textbf{Online ST} has very limited performance due to continuously fitting on self-errors, where it even performs worse than Baseline when labels are extremely scarce (1/50). We also observe that its performance remains similar without data augmentation on the unlabeled subset, echos our analysis that consistency regularization cannot be integrated for online self-training without multiple forward passes (Section \ref{sec:22}).

\textbf{CBST} is an offline method, thus it is less sensitive to self-error than Online ST and consistently improves over Baseline by a clear margin. However, it only conducts self-training without considering pseudo label noise. In our experiments, performance increase stops at iteration three for CBST while DMT benefits from all 5 iterations. Because pseudo label noise prevents further improvements on CBST when using more unlabeled data.

\begin{figure}[t]
\centering
\includegraphics[scale=0.9]{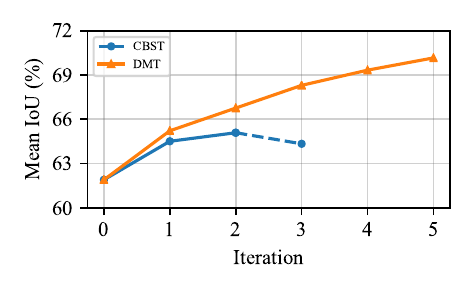}
\caption{Comparison of DMT and CBST (both COCO initialized) on one PASCAL VOC 2012 1/20 split. Iteration 0 represents training on the labeled subset only. CBST performance starts to degrade at iteration three due to too much pseudo label noise. }
\label{fig6}
\end{figure}

\textbf{DST} requires half the computing budget compared to DMT and performs well. As fine-tuning goes on in each iteration, a relatively large learning rate and data augmentations drive the model to deviate from its previous-self\footnote{Previous-self denotes the model state that produced the pseudo labels before the current mutual training iteration.} rapidly. Thus, sufficient inter-model disagreement is provided for dynamic weighting to take effects. While in DMT, larger inter-model disagreement by starting from different model initializations naturally enables better final results, especially on the more challenging 1/50 split.

\textbf{DMT-Naive} is a simpler formulation to integrate inter-model disagreement, the performance is even comparable to DMT on the 1/20 split. But its performance degrades significantly when label noise is severe (1/50 split). Although this naive policy outperforms CBST on this task, its performance is similar or worse than CL on CIFAR-10, indicates poor generalization ability.

\textbf{DMT-Flip} is more complex than DMT. But its performance is similar to DMT, differences are at the level of random variations. We suspect by flipping labels, there is a similar drawback as online self-training: fitting newly made self-errors (Section \ref{sec:3}). Thus the results are no better than ignoring those labels (usually fewer than $5\%$ in training). Moreover, considering simplifying the policy also brings notable degradation (DMT-Naive), the three-cases setup in DMT (Eq. \ref{eq:2}) offers a reasonable trade-off between performance and complexity.

\section{Conclusions and Discussions}
\label{sec:7}
In this paper, we have proposed Dynamic Mutual Training (DMT) to counter the pseudo supervision noise by a re-weighted loss function based on the inter-model disagreement. Furthermore, we have adapted DMT to an iterative framework for better performance in both image classification and semantic segmentation. DMT is flexible and easy to implement.
We have evaluated the proposed method on different datasets including CIFAR-10, PASCAL VOC 2012 and Cityscapes.
The experiments (comparisons and ablations) clearly demonstrate the effectiveness of the proposed DMT, and show its state-of-the-art outcomes in classification and segmentation.

We find that DMT is more promising in semantic segmentation than image classification, probably because dynamic weighting exploits pixels with high-quality pseudo labels in an image and provides better pseudo supervision on each image overall. In addition, image classification on CIFAR-10 requires re-training for each iteration and the two models do not have equal classification ability throughout training, thus making it difficult to exploit inter-model disagreement. Besides, it is hard to estimate confidence when recent image classification models require heavy data augmentation in training, confidence distribution is quite different compared to the generated pseudo labels. Thus, a better confidence estimation process could bring further gains, e.g. multiple forward pass statistics (at the cost of computing). Also advances in the learning with noisy labels community may be potentially useful, which is worthy of future investigations.

Our work shares the common limitation of most offline self-training methods: the initial model learned on the labeled subset may be insufficient if the labels are too few, e.g. 100 labels on Cityscapes. Better pre-trained weights, e.g. COCO pre-trained weights for PASCAL VOC 2012, can potentially alleviate this issue, given a small gap to Oracle on PASCAL VOC 2012 ($11.71\%$, Tab. \ref{table:pascal}). If off-the-shelf pre-trained weights are unavailable, self-supervised learning \cite{zhai2019s4l} is a good way to initiate learning. We intend to investigate how self-supervised learning can help semi-supervised learning in the future, especially for structured tasks like semantic segmentation.

\section{Acknowledgements}

This work was partially supported by National Key Research and Development Program of China (No. 2019YFC1521104), Art major project of National Social Science Fund (I8ZD22). The author Qianyu Zhou is supported by Wu Wenjun Honorary Doctoral Scholarship, AI Institute, Shanghai Jiao Tong University. Also, the authors would like to thank Shikun Liu (Imperial College London) for insightful discussions and identifying an incorrect data augmentation policy in the earlier semantic segmentation code.

\appendix

\section{Extra Qualitative Results}
\label{sec:a1}

\begin{figure}[t]
\centering
\includegraphics[scale=0.6]{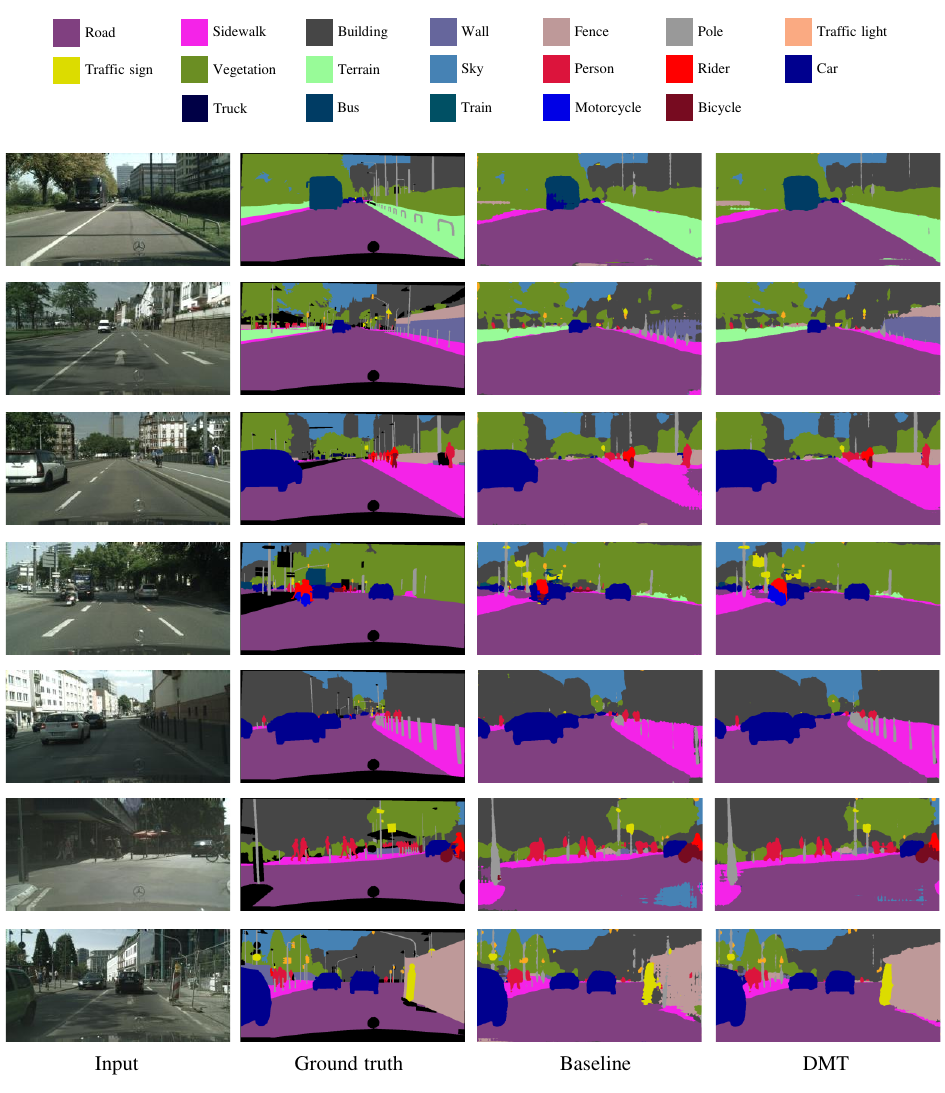}
\caption{Qualitative results on Cityscapes. Models are trained with 1/8 pixel-level labeled data in the training set. Black regions are ignored in ground truth. }
\label{fig11}
\end{figure}

Qualitative comparisons among Baseline (supervised only), our method (DMT) and ground truth on Cityscapes are shown in Fig. \ref{fig11}. Specifically, there is confusion among similar classes from Baseline (column 3), such as \textit{bus} and \textit{car} (row 1), \textit{wall}, \textit{fence} and \textit{building} (row 2), \textit{road} and \textit{sidewalk} (row 3). In contrast, DMT does a better job (column 4). DMT also detects the existence of small objects better, e.g. \textit{motorcycle} in row 4, \textit{pole} in row 5. While Baseline claims existence of non-existent class \textit{sky} in row 6. In row 7, where the real-world \textit{fence} is more complex, Baseline produces chaotic results which become consistent after dynamic mutual training.

\section{Proper Baseline Training}
\label{sec:a3}

A concern of non-fully training baselines in semi-supervised learning has been raised in \cite{oliver2018realistic}. We find that other than unifying data augmentation schemes and tricks (e.g. using the same strong augmentations and mixup in CIFAR-10 baselines), one important factor is the number of epochs. For example, we have a labeled subset that is 1/8 the entire dataset, 8 times more epochs (i.e. keep the number of steps unchanged) are too many, and the same number of epochs are too few. Thus, we make a compromise between them and train for $\sqrt{\frac{1}{labeled~ratio}} N$ epochs, where $N$ is the number of epochs used in Oracle training, which is 300, 30, 60 for CIFAR-10, PASCAL VOC 2012 and Cityscapes, respectively. As a result, the reported baseline performance in this paper is noticeably higher than previous works, sometimes even comparable to some previous state-of-the-art methods on Cityscapes (Tab. 5).

In our fully-supervised baselines, the learning rate is set to $0.2$ (CIFAR-10), $2 \times 10^{-3}$ (PASCAL VOC 2012), and $4 \times 10^{-3}$ (Cityscapes).

\section{Realistic Validation}
\label{sec:a4}

Oliver et al. \cite{oliver2018realistic} has raised a concern of using unrealistically large validation sets in semi-supervised image classification. For instance, most prior arts split a validation set of 5,000 images for hyper-parameter tuning on CIFAR-10, even larger than their labeled subset. While we use only 200 images (we call this validation set \textit{valtiny} and will release it along with our source codes). Intuitively, 200 images can hardly tell the difference between two models with $100\% / 200 = 0.5\%$ accuracy gap. Different from semantic segmentation, since the mean IoU is a fine-grained metric that can work with a small validation set.

\textbf{Fine-grained testing.} To test with fewer images, we propose fine-grained testing for image classification, where we count the probability of being correct. Concretely, if a model makes the right decision on an image, we count it as the probability that model assigns for the correct class instead of 1. In this way, we assess not only whether a prediction is correct, but also how correct it is. We observe it helpful when normal testing can not tell the difference, especially when comparing similar setups (a few different hyper-parameter values). But fine-grained testing does not work well when comparing mixup methods and non-mixup methods, since mixup is better class-calibrated \cite{thulasidasan2019mixup}.

\textbf{Hyper-parameter choice on $\gamma$.} $\gamma$ are viewed as hyper-parameters, similar to other common hyper-parameters, such as the number of epochs, learning rate, weight decay, etc.  We choose the optimal gamma values using a small-ranged grid-search method, which is a typical technique for hyper-parameter tuning. We choose a single gamma value for each dataset, which has stable performance across different labeled ratios. The trial grid for our grid-search method was: (1, 3, 5, 7, 9) on PASCAL VOC, which validated the best gamma value 5, then 4 and 6 is tried but no obvious improvement is observed. It is further tuned for the other datasets: (3, 4, 5, 6) for CIFAR and Cityscapes until no notable improvement can be observed. To be realistic [35], we used a small validation set for each dataset, as well as empirical observations (e.g. loss curve, intuitions from other tasks) %

\section{Complexity}
\label{a:r4}

Firstly, our method does not contain any modification to the model architecture. Thus in test phase, it has the same computational complexity as the supervised baseline model. The only difference is model training, and  details are shown in following Table \ref{table:1}.

\begin{table}[h]
\caption{Training time of DMT, measured by a single 2080 Ti GPU.}
\label{table:1}
\centering
\begin{tabular}{ccc} \toprule
Dataset & Training time (hour) & Relative time to Oracle training \\
\midrule

PASCAL VOC 2012 & 8 & 2.5x \\
Cityscapes & 6 & 2.5x \\
CIFAR-10 & 10 & 8x \\

\bottomrule
\end{tabular}
\end{table}

In segmentation, 5 iterations of DMT (including pseudo label generation) plus baseline training, take only roughly 2.5x the time of training a fully-supervised model (Oracle), because each iteration is fine-tuning and thus very fast. Considering two models are produced at the end, the training cost is similar to Oracle training.

In classification, the models are re-trained 5 times, and each time it takes longer to converge than the fully-supervised baseline (750 epochs, compared to 300 epochs for Oracle). The overall training time seems much longer. We use the same training epochs as CL \cite{cascante2020curriculum}. As noted in CL \cite{cascante2020curriculum}, this is still much faster than other semi-supervised learning methods on CIFAR-10.

\section{Extra Illustrations}

\begin{figure}[h]
\centering
\includegraphics[scale=1.2]{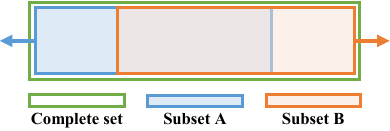}
\caption{Difference maximized sampling. The complete set is randomly shuffled at first, and subset A and B are drawn with an equal size but with fewest overlapped samples.}
\label{fig:r2}
\end{figure}

\bibliographystyle{elsarticle-num-names} 
\bibliography{full}

\end{document}